\documentclass{article}

\usepackage[accepted]{icml2012}

\usepackage{times}

\usepackage{bm}
\usepackage{graphicx}
\usepackage{latexsym}
\usepackage{amsmath,amsfonts,amssymb,mathrsfs,color}
\usepackage{algorithm,algorithmic}
\usepackage{latexsym}
\usepackage{subfigure}
\usepackage{pifont}
\usepackage{amsthm}

\input{commands}

\newtheorem{theorem}{Theorem}[section]
\newtheorem{lemma}[theorem]{Lemma}

\newtheorem{corollary}[theorem]{Corollary}
\newtheorem{proposition}[theorem]{Proposition}

\newenvironment{definition}[1][Definition]{\begin{trivlist}\item[\hskip \labelsep {\bfseries #1}]}{\end{trivlist}}

\newenvironment{remark}[1][Remark]{\begin{trivlist}
\item[\hskip \labelsep {\bfseries #1}]}{\end{trivlist}}

\newcommand{\SongEmbeddings}{DBLP:conf/icml/SongHSF09}

\newcommand{\Id}{{\rm Id \,}}
\newcommand{\hatmu}{\hat \mu}
\newcommand{\Err}{{\mathcal{E}}}
\newcommand{\hatErr}{\what{\Err}}

\newcommand{\G}{\Gamma }

\renewcommand{\loss}{{\rm loss}}

\begin{document}
\twocolumn[
\icmltitle{Conditional mean embeddings as regressors}
\icmlauthor{Steffen Gr\"unew\"alder\footnotemark[1]}{steffen@cs.ucl.ac.uk}
\icmlauthor{Guy Lever\footnotemark[1]}{g.lever@cs.ucl.ac.uk}
\icmlauthor{Luca Baldassarre}{l.baldassarre@cs.ucl.ac.uk}
\icmlauthor{Sam Patterson$^{*}$}{sam.x.patterson@gmail.com}
\icmlauthor{Arthur Gretton$^{*,\dagger}$}{arthur.gretton@gmail.com}
\icmlauthor{Massimilano Pontil}{m.pontil@cs.ucl.ac.uk}
\icmladdress{CSML and $^*$Gatsby Unit, University College London, UK, $^\dagger$ MPI for Intelligent Systems}

\icmlkeywords{Conditional Mean Embedding, Vector valued RKHS, Regression, Minimax rates, Sparsity, Cross Validation}

\vskip 0.3in
]

\footnotetext[1]{\textbf{Equal contribution.}  -- Supplementary on arXiv.}

\begin{abstract}
We demonstrate an equivalence between reproducing kernel Hilbert space (RKHS) embeddings of conditional distributions and vector-valued regressors. This connection introduces a natural regularized loss function which the RKHS embeddings minimise, providing an intuitive understanding of the embeddings and a justification for their use. Furthermore, the equivalence allows the application of vector-valued regression methods and results to the problem 
of learning conditional distributions.
Using this link we derive a sparse version of the embedding by considering  alternative formulations. Further, by applying convergence results for vector-valued regression to the embedding problem we derive minimax convergence rates which are  $O(\log(n)/n)$ -- compared to current state of the art rates of $O(n^{-1/4})$ -- and are valid under milder and more intuitive assumptions. These minimax upper rates coincide with lower rates up to a logarithmic factor, showing that the embedding method achieves nearly optimal rates.
We study our sparse embedding algorithm in a reinforcement learning task where the algorithm shows significant improvement in sparsity over an incomplete Cholesky decomposition.
\end{abstract}

\section{Introduction/Motivation}

In recent years a framework for embedding probability distributions into
reproducing kernel Hilbert spaces (RKHS) 
has become increasingly popular \citep{Smola07}. One example of this theme has been the representation of conditional 
expectation operators as RKHS functions, known as \emph{conditional mean embeddings} \citep{\SongEmbeddings}. Conditional expectations appear naturally in many machine learning tasks, and the RKHS representation  of such expectations has two important advantages:
 first, conditional mean embeddings do not require solving difficult intermediate problems such as density estimation and numerical integration;  and second, these embeddings may be used to compute conditional expectations   directly on the basis of observed samples. Conditional mean embeddings have been successfully applied to inference in graphical models, reinforcement learning, subspace selection, and conditional independence testing \citep{FukGreSunSch08,FukBacJor09,DBLP:conf/icml/SongHSF09,DBLP:journals/jmlr/SongGG10,RLEmbNoStar}. 



The main motivation for conditional means in Hilbert spaces has been
to generalize the notion of conditional expectations from finite
cases (multivariate Gaussians, conditional probability tables, and so on).
 Results have been established for the
convergence of these embeddings in RKHS norm \citep{DBLP:conf/icml/SongHSF09,DBLP:journals/jmlr/SongGG10},
which show that conditional mean embeddings behave
in the way we would hope (i.e., they may be used in obtaining conditional
expectations as inner products in feature space, and these estimates are consistent under 
smoothness conditions). 
Despite these valuable results, the
characterization of conditional mean embeddings remains incomplete, since these embeddings
have not  been
 defined in terms of the optimizer of a given {\em loss function}. 
This makes it difficult to extend these results, and has hindered the use of standard 
techniques like cross-validation for parameter estimation.

In this paper, we demonstrate that the conditional mean embedding is the
solution of a vector-valued regression problem with a natural loss,
 resembling the standard Tikhonov regularized least-squares problem
in multiple dimensions. Through this link, it is possible to access the
rich theory of vector-valued regression
\citep{MIC05,CarDevToi06,CAP07,CAP08}.
 We
demonstrate the utility of this connection by providing novel
characterizations of conditional mean embeddings, with important theoretical
and practical implications. 
On the theoretical side, we establish novel convergence results for
RKHS embeddings, giving a significant improvement over the rate
of $O(n^{-1/4})$ due to \citet{DBLP:conf/icml/SongHSF09,DBLP:journals/jmlr/SongGG10}. 
We derive a
faster $O(\log(n)/n)$ rate which holds over large classes of
probability measures, and requires milder and more intuitive
assumptions. We also show 
our rates are optimal up to a $\log(n)$ term, following the analysis of \citet{CAP07}.
On the practical side, we derive an alternative sparse version of the
embeddings which resembles the Lasso method, and provide  a
cross-validation scheme for parameter selection.

\section{Background}

In this section, we recall some background results concerning RKHS embeddings and vector-valued RKHS. For an introduction to scalar-valued RKHS we refer the reader to \citep{BER04}.



\subsection{Conditional mean embeddings} \label{EmbeddingsIntro}
Given sets $\cX$ and $\cY$, with a distribution $P$ over random variables $(X,Y)$ from $\cX\times\cY$ we consider the problem of learning expectation operators corresponding to the conditional distributions $P(Y|X=x)$ on $\cY$ after conditioning on $x\in\cX$. Specifically, we begin with a kernel $L:\cY\times\cY\into\R$, with corresponding RKHS $\cH_L\subseteq \R^\cY$,  
and study the problem of learning, for every $x \in \cX$, the {\em conditional expectation mapping} $\cH_L \ni h\mapsto \Expect [h(Y)|X=x]$. 
Each such map can be represented as 
\begin{align}
\Expect [h(Y)|X=x] = \lang h , \mu(x) \rang_L, \nn
\end{align}
where the element $\mu(x)\in\cH_L$ is called the \emph{(conditional) mean embedding} of $P(Y|X=x)$.
Note that, for every $x$, $\mu(x)$ is a function on $\cY$. It is thus apparent that $\mu$ is a mapping from $\cX$ to $\cH_L$, a point which we will expand upon shortly.

We are interested in the problem of estimating the embeddings $\mu(x)$ given an i.i.d. sample $\{ (x_i,y_i) \}_{i=1}^n$ drawn from $P^n$. Following \citep{\SongEmbeddings,DBLP:journals/jmlr/SongGG10}, we  define a second kernel $K:\cX\times\cX\into \R$ with associated RKHS $\cH_K$, and consider the estimate
\begin{align}
\hatmu(x) :=  \sum_{i=1}^n \alpha_i(x) L(y_i,\cdot), 
\label{EmbeddingExpansion}
\end{align}
where $\alpha_i(x) = \sum_{j=1}^n W_{ij} K(x_j,x)$, and where $\bW:=(\bK + \lambda n \bI)^{-1}$, $\bK = (K(x_i,x_j))_{ij=1}^n$, and $\lambda$ is a chosen regularization parameter.
This expression suggests that the conditional mean embedding is the solution to an underlying regression problem: we will formalize this link in Section \ref{sec:estimatingCondExp}. In the remainder of the present section, we introduce the necessary terminology and theory for vector valued regression in RHKSs.

\subsection{Vector-valued regression and RKHSs} \label{RegressionIntro}

We recall some background on learning vector-valued functions using kernel methods \citep[see][for more detail]{MIC05}. We are given a sample $\{(x_i,v_i)\}_{i \leq m}$ drawn i.i.d. from some distribution over $\cX\times\cV$, where $\cX$  is a non-empty set and 
  $(\cV,\lang \cdot , \cdot \rang_\cV)$ is a Hilbert space.
Our goal is to find a function $f:\cX\to\cV$ with low error, as measured by
\begin{align}
\Expect_{(X,V)}[||f(X) - V||_\cV^2]. \label{TrueError}
\end{align}
This is the \emph{vector-valued regression} problem (square loss).

One approach to the vector-valued regression problem is to model the regression function as being in a vector-valued RKHS of functions  taking values in $\cV$, which can be defined by analogy with the scalar valued case.
\begin{definition}
A Hilbert space $(\cH,\langle \cdot, \cdot \rangle_\G)$ of functions $h: \cX \rightarrow \cV$ is an RKHS if for all $x \in \cX, v \in \cV$ the linear functional $h \mapsto \langle v, h(x) \rangle_\cV$ is continuous.  
\end{definition}


The reproducing property for vector-valued RHKSs follows from this definition \citep[see][Sect. 2]{MIC05}.  By the Riesz representation theorem, for each $x\in\cX$ and $v\in\cV$, there exists a linear operator from $\cV$ to $\cH_\G$ written $\G_xv\in\cH_\G$,  such that for all $h\in\cH_\G$,
$$
\lang v , h(x)  \rang_\cV = \lang h , \G_x  v \rang_\G. \nn
$$
It is instructive to compare to the scalar-valued RKHS $\cH_K$, for which the linear operator of evaluation $\delta_x$ mapping $h\in\cH_K$ to $h(x)\in\R$  is continuous: then Riesz implies there exists a $K_x$ such that $yh(x)=\langle h,yK_x\rangle_K$.

We next introduce the vector-valued reproducing kernel, and show its relation to $\G_x$. Writing as $\mathcal{L}(\cV)$ the space of bounded linear operators from $\cV$ to $\cV$, the reproducing kernel $\G(x,x')\in \mathcal{L}(\cV)$ is defined as
\begin{align}
\G(x,x')v = (\G_{x'}v) (x) \in \cV. \nn
\end{align}

From this definition and the reproducing property, the following holds \citep[Prop. 2.1]{MIC05}.

\begin{proposition}\label{prop:vectorKernelDefinition}
A function $\G: \cX \times \cX \rightarrow \mathcal{L}(\cV)$ is a kernel if it satisfies: (i) $\G(x,x') = \G(x',x)^*$, (ii) for all $n \in \mathbb{N}, \{(x_i,v_i)\}_{i \leq n} \subseteq \cX \times \cV$  we have that 
$ \sum_{i,j \leq n} \langle v_i, \G(x_i,x_j) v_j \rangle_\cV \geq 0$.
\end{proposition}
It is again helpful to consider the scalar case: here, $\langle K_x,K_{x'}\rangle_K=K(x,x')$, and
to every positive definite kernel $K(x,x')$ there corresponds a unique (up to isometry) RKHS for which $K$ is the reproducing kernel. Similarly,  if $\G: \cX \times \cX \rightarrow \mathcal{L}(\cV)$ is a kernel in the sense of Proposition \ref{prop:vectorKernelDefinition},  there exists a unique (up to isometry) RKHS, with $\G$ as its reproducing kernel \citep[Th. 2.1]{MIC05}. 
Furthermore, the RKHS $\cH_\G$ can be described as the RKHS limit of finite sums; that is, $\cH_\G$ is up to isometry equal to the closure of the linear span of the set 
$\left\{\G_{x} v : x \in \cX, v \in \cV \right\}$, wrt the RKHS norm $\| \cdot \|_\G$.

Importantly, it is possible to perform regression in this setting. One approach to the vector-valued regression problem is to replace the unknown true error (\ref{TrueError}) with a sample-based estimate $\sum_{i=1}^n ||v_i - f(x_i)||_\cV^2$,  restricting $f$ to be an element of an RKHS $\cH_\G$ (of vector-valued functions), and regularizing w.r.t. the $\cH_\G$ norm, to prevent overfitting.
\ We  thus arrive at the following regularized empirical risk,
\begin{align}
\hatErr_{\lambda}(f) := 
\sum_{i=1}^n ||v_i - f(x_i)||_\cV^2 + \lambda ||f||_\G^2. \label{VVRegressionProblem}
\end{align}
\begin{theorem}\label{VVTheorem}\citep[][Th. 4]{MIC05}
If $f^*$ minimises $\hatErr_{\lambda}$ in $\cH_\G$ then it is unique and has the form,
$$
f^* = \sum_{i=1}^n \G_{x_i} c_i, \nn
$$
where the coefficients $\{c_i\}_{i \leq n}$, $c_i \in \cV$ are the unique solution of the system of linear 
equations
$$
\sum_{i\leq n} (\G(x_j,x_i) + \lambda \delta_{ji})c_i = v_j, \quad 1 \leq j \leq n. \nn
$$  
\end{theorem}


In the scalar case we have that $|f(x)| \leq \sqrt{K(x,x)} \|f\|_K $.
Similarly it holds that  $\|f(x) \| \leq |||\G(x,x)|||~\|f\|_\G$, where 
$|||\cdot|||$ denotes the operator norm \citep[][Prop. 1]{MIC05}. Hence, 
if $|||\G(x,x)||| \leq B $ for all $x$ then 
\begin{equation}
 \|f(x)\|_\cV \leq B \|f\|_\G.
\label{cor:RKHSDom}
\end{equation}


Finally, we need a result that tells us when all functions in an RKHS are continuous. In the scalar case this is guaranteed if $K(x,\cdot)$ is continuous for all $x$ and 
$K$ is bounded. In our case we have \citep{CarDevToi06}[Prop. 12]:
\begin{corollary} \label{Cor:ContRKHS}
If $\cX$ is a Polish space, $\cV$ a separabe Hilbert space and the mapping $x \mapsto \G(\cdot,x)$ is continuous, then $\cH_\G$ is a subset of the set of continuous functions from $\cX$ to $\cV$.
\end{corollary}
\section{Estimating conditional expectations}\label{sec:estimatingCondExp}

In this section, we show the problem of learning conditional mean embeddings can be naturally formalised in the framework of vector-valued regression, and in doing so we derive an equivalence between the conditional mean embeddings and a vector-valued regressor.

\subsection{The equivalence between conditional mean embeddings and a vector-valued regressor} \label{Equivalence}

 Conditional expectations $\Expect[h(Y)|X=x]$ are linear in the argument $h$ so that, when we consider $h\in\cH_L$, the Riesz representation theorem implies the existence of an element $\mu(x)\in\cH_L$ such that $\Expect[h(Y)|X=x] = \langle h, \mu(x) \rangle_L$ for all $h$. That being said,  the dependence of $\mu$ on $x$ may be complicated. A natural optimisation problem associated to this approximation problem is to therefore find a function $\mu:\cX \into \cH_L$ such that the following objective is small
\begin{align}
\hspace{-2mm}\Err[\mu] := \sup_{\|h\|_L \leq 1}\Expect_{X}\left[\left( \Expect_Y[h(Y)|X] - \lang h, \mu(X) \rang_L \right)^2 \right]. \label{Objective}
\end{align}

Note that the risk function cannot be used directly for estimation, because we do not observe $\Expect_{Y}[h(Y)|X]$, but rather pairs $(X,Y)$ drawn from $P$. However, we can bound this risk function with a surrogate risk function that has a sample based version, 
\begin{align}
&\sup_{\|h\|_L \leq 1}\Expect_{X}\left[\left( \Expect_Y[h(Y)|X] - \lang h , \mu(X) \rang_L \right)^2 \right] \nn\\ 
&~~~=  \sup_{\|h\|_L \leq 1}\Expect_{X}\left[\left( \Expect_Y[\langle h, L(Y,\cdot) \rangle_L |X] - \lang h , \mu(X) \rang_L \right)^2 \right] \nn\\ 
&~~~\le \sup_{\|h\|_L \leq 1}\Expect_{X,Y}\left[ \lang h, L(Y,\cdot) - \mu(X) \rang_L^2 \right] \nn\\
&~~~\le \sup_{\|h\|_L \leq 1} \|h\|_L^2 \, \Expect_{X,Y}\left[ \| L(Y,\cdot) - \mu(X)\|_L^2 \right] \nn\\
&~~~= \Expect_{(X,Y)}\left[ || L(Y,\cdot) - \mu(X)||_L^2  \right],\label{Dual}
\end{align}
where the first and second bounds follow by Jensen's and 
Cauchy-Schwarz's inequalities, respectively. Let us denote this surrogate risk function as
\begin{align}
\Err_s[\mu] := \Expect_{(X,Y)}\left[ || L(Y,\cdot) - \mu(X)||_L^2  \right]. \label{Surrogate}
\end{align}
The two risk functions $\Err$ and $\Err_s$ are closely related and in Section~\ref{RelatingObjectives} we examine their relation.
 
We now replace the expectation in (\ref{Dual}) with an empirical estimate, to obtain the sample-based loss,
\begin{align}
\hatErr_n[\mu] := \sum_{i=1}^n || L(y_i,\cdot) - \mu(x_i) ||_L^2. \label{Loss}
\end{align}
Taking (\ref{Loss}) as our empirical loss, then following Section~\ref{RegressionIntro} we add a regularization term to provide a well-posed problem and prevent overfitting,
\begin{align}
\hatErr_{\lambda,n}[\mu] := \sum_{i=1}^n || L(y_i,\cdot) - \mu(x_i) ||_L^2 + \lambda ||\mu||_{\G}^2.\label{RegularizedLoss}
\end{align}
We denote the minimizer of (\ref{RegularizedLoss}) by $\hat \mu_{\lambda,n}$,
\begin{equation}
\hat \mu_{\lambda,n} := \argmin_{\mu} \left\{ \hatErr_{\lambda,n}[\mu] \right\}. \label{Problem}
\end{equation}
Thus, recalling (\ref{VVRegressionProblem}), we can see that 
the problem (\ref{Problem}) is posed as a vector-valued regression problem with the training data now considered as $\{(x_i,L(y_i,\cdot))\}_{i=1}^n$ (and we identify $\cH_L$ with the general Hilbert space $\cV$ of Section~\ref{RegressionIntro}). From Theorem~\ref{VVTheorem}, the solution is
\begin{equation}
\hat \mu_{\lambda,n} = \sum_{i=1}^n \G_{x_i} c_i, \label{MuForm}
\end{equation}
where the coefficients $\{c_i\}_{i \leq n}$, $c_i \in \cH_L$ are the unique solution of the linear 
equations
$$
\sum_{i\leq n} (\G(x_j,x_i) + \lambda \delta_{ji})c_i = L(y_j,\cdot), \quad 1 \leq j \leq n. \nn
$$
It remains to choose the kernel $\G$. Given a real-valued kernel $K$ on $\cX$, a natural choice for the RKHS $\cH_{\G}$ would be the space of functions from $\cX$ to $\cH_L$ whose elements are defined as functions via $(h,K(x,\cdot))(x') := K(x,x')h$, which is isomorphic to $\cH_L \otimes \cH_K$, with inner product 
\begin{align}
\lang g K(x,\cdot) , h K(x',\cdot) \rang_{\G} := \lang g  , h  \rang_{L} K(x,x')  
\end{align}
for all $g,h\in\cH_L$. Its easy to check that this satisfies the conditions to be a vector-valued RKHS
-- in fact it corresponds to the choice $\G(x,x') = K(x,x')\Id$, where $\Id:\cH_L \to\cH_L$ is the identity map on $\cH_L$. The solution to (\ref{Problem}) with this choice is then given by (\ref{MuForm}), with
\begin{align}
\sum_{i\leq n} (K(x_j,x_i) + \lambda \delta_{ji})c_i = L(y_j,\cdot)&, \quad 1 \leq j \leq n \nn\\
c_i =  \sum_{j\leq n} W_{ij} L(y_j,\cdot)&, \quad 1 \leq i \leq n,\nn
\end{align}
where $\bW = (\bK + \lambda \bI)^{-1}$, which corresponds exactly the embeddings (\ref{EmbeddingExpansion}) presented in \citep{DBLP:conf/icml/SongHSF09,DBLP:journals/jmlr/SongGG10} (after a rescaling of $\lambda$). Thus we have shown that the embeddings of Song et al. are the solution to a  regression problem for a particular choice of operator-valued kernel. Further, the loss defined by (\ref{Surrogate}) is a key error functional in this context since it is  the objective which the estimated embeddings attempt to minimise. In Sec.~\ref{RelatingObjectives} we will see that this does not always coincide with (\ref{Objective}) which may be a more natural choice. In Sec.~\ref{Rates} we  analyze the performance of the embeddings defined by (\ref{Problem}) at minimizing the objective (\ref{Surrogate}).

\subsection{Some consequences of this equivalence}

We derive some immediate benefits from the connection described above. Since the embedding problem has been identified as a regression problem with training set $\cD := \{ (x_i,L(y_i,\cdot))\}_{i=1}^m$, we can define a cross validation scheme for parameter selection in the usual way: by holding out a subsample $\cD_{{\rm val}} = \{ (x_{t_i},L(y_{t_j},\cdot))\}_{j=1}^{T} \subset \cD$, we can train embeddings $\hat \mu$ on $\cD\backslash \cD_{{\rm val}}$ over a grid of kernel or regularization parameters, choosing the parameters achieving the best error $\sum_{j=1}^{T} || \hat \mu (x_{t_j}) - L(y_{t_j},\cdot) ||_L^2$ on the validation set (or over many folds of cross validation). Another key benefit will be a much improved performance analysis of the embeddings, presented in Section~\ref{Rates}.


\begin{table}[t]
\begin{tabular}{|l|l|}
\hline 
Input/Output space &  (i) $\cX$ is Polish. \\
& (ii) $\cV$ is separable.  \textit{(f.b.a.)}\\
& (iii) $\exists C> 0$ such that $\forall x \in \cX$ \\
& \quad \enspace \enspace $\text{Tr}(\G(x,x)) \leq C$  holds. \\
\hline 
Space of regressors & (iv) $\cH_\G$ is separable. \\
& (v) All $\Gamma_x^*$ are HS. \textit{(f.b.a.)} \\
& (vi) $B:(x,y) \rightarrow  \langle  f, \G(y,x) g \rangle_\cV $ \\
& \quad \enspace \enspace is measurable $\forall f, g \in \cV$. \\
\hline 
True distribution & (vii) $L(y,y) < \infty$ for all $y \in \cY$. \\
& (viii) $ \mathcal{E}_s[\mu^*] = \inf_{\mu \in \cH_\G} \mathcal{E}_s[\mu]$. \\
\hline 
\end{tabular}
\caption{Assumptions for Corollary \ref{Cor:CapUpper} and \ref{Cor:CapLower}. f.b.a. stands for fulfilled by assumption that $\cV$ is finite dimensional.}
\label{tab:Ass}
\end{table}

\subsection{Relations between the error functionals $\Err$ and $\Err_s$} \label{RelatingObjectives}

In Section~\ref{Equivalence} we introduced an alternative risk function $\Err_s$ for $\Err$, which we used to derive an estimation scheme to recover conditional mean embeddings. We now examine the relationship between the two risk functionals. When the true conditional expectation on functions $h \in \cH_L$ 
can be represented through  an element $\mu^* \in \cH_\G$ then $\mu^*$ minimises both objectives.
\begin{theorem}[Proof in App. \ref{sec:MinEqual}] \label{thm:MinEqual}
If there exists a $\mu^* \in \cH_\G$ such that for any $h\in \cH_L$: 
$\Expect[h |X] = \lang h,  \mu^*(X) \rang_L$ $P_\cX$-a.s., then $\mu^*$ is the 
$P_\cX$-a.s. unique minimiser of both objectives:
\begin{align*}
& \mu^* = \argmin_{\mu \in \cH_\G} \Err[\mu] =  \argmin_{\mu \in \cH_\G} \Err_s[\mu]
\quad P_\cX \, a.s.
\end{align*}
\end{theorem} 
Thus in this case, the embeddings of \citet[e.g.][]{\SongEmbeddings} minimise both (\ref{Objective}) and (\ref{Surrogate}). More generally, however, this may not be the case. Let us define an element $\hat \mu$ that is $\delta$ close w.r.t. the error $\Err_s$ to the minimizer $\mu'$ of $\Err_s$ in $\cH_\G$ (this might for instance be the minimizer of the empirical regularized loss for sufficiently many samples).
We are interested in finding conditions under which $\Err(\hat \mu)$ is not much worse than a good {\em approximation} $\mu^*$ in $\cH_\G$ to the conditional expectation. The sense in which $\mu^*$ approximates the conditional expectation is somewhat subtle: $\mu^*$ must closely approximate the conditional expectation of functions $\mu \in \cH_\G$ under the original loss $\Err$ (note that the loss $\Err$ was originally defined in terms of functions $h\in\cH_L$).

\begin{theorem}[Proof in App. \ref{sec:MinEqual}] \label{thm:bound}
Let $\mu'$ be a minimiser  of $\Err_s$  and $\hat \mu$ be an element of $\cH_\G$ with  $\Err_s[\hat \mu] \leq \Err_s[\mu'] + \delta$. 
Define, $ \mathcal{A} := \{ (\eta,\tilde \mu)  \, | \, \eta^2 = \sup_{||\mu||_\G \leq 1}\Expect_X \left[ \Expect_Y[ \mu(X) |X]  -  \lang \mu(X) ,  \tilde \mu(X) \rang_L\right]^2  \}$
, then
\begin{align*}
&\Err[\hat \mu]   \leq 
\inf_{(\eta,\mu^*) \in \mathcal{A}}
\left(\sqrt{\Err[\mu^*]} + \sqrt{8\eta ( \|\mu^*\|_\G 
+\|\hat \mu\|_\G)} + \delta^{\frac{1}{2}}\right)^2_.
\end{align*}
\end{theorem}

Apart from the more obvious condition that $\delta$ be small, the above theorem suggests that  $||\hat \mu||_\G$ should also be made small for the solution $\hat \mu$ to have low error $\Err$.
 In other words, even in the infinite sample case, the regularization of $\hat \mu$ in  $\cH_\G$ is important.

\section{Better convergence rates for embeddings} \label{Rates}
\label{sec:ConvRate}
The interpretation of the mean embedding as a vector valued regression problem allows us to 
apply regression minimax theorems to study convergence rates of the embedding estimator. These rates are considerably better than the current state of the art for the embeddings, and hold under milder and more intuitive assumptions.

We start by comparing the statements which we derive from \citep[][Thm.s 1 and 2]{CAP07} with the known convergence results for the embedding estimator. We follow this up with a discussion of the rates and a comparison of the assumptions.

\subsection{Convergence theorems}
We  address the performance of the embeddings defined by (\ref{Problem}) in terms of asymptotic guarantees on the loss $\Err_s$ defined by (\ref{Surrogate}). \citet{CAP07} study uniform convergence rates for regression. Convergence rates of learning algorithms can not be uniform on the set of all probability distributions if the output vector space is an infinite dimensional RKHS \cite{CAP07}[p. 4]. It is therefore necessary to restrict ourselves to a subset of probability measures. This is done by \citet{CAP07} by defining families of probability measures $\mathscr{P}(b,c)$  indexed by two parameters 
$b \in ]1,\infty]$ and $c \in [1,2]$.
We discuss the family $\mathscr{P}(b,c)$ in detail below. The important point at the moment is that $b$ and $c$ affect the optimal schedule for the regulariser $\lambda$ and the convergence rate. The rate of convergence is better for higher $b$ and $c$ values. \citet{CAP07} provide convergence rates for all choices of $b$ and $c$. We restrict ourself to the best case $b = \infty, c>1$ and the worst case\footnote{Strictly speaking the worst case is $b \downarrow 1$ (see supp.).} $b = 2, c= 1$.

We recall that the estimated conditional mean embeddings $\hat \mu_{\lambda,n}$ are given by (\ref{Problem}), where $\lambda$ is a chosen regularization parameter. We assume $\lambda_n$ is chosen to follow a specific schedule, dependent upon $n$: we denote by $\hat \mu_{n}$ the embeddings following this  schedule and $\mu':=\argmin_{\mu\in\cH_\G} \Err_s[\mu]$. Given this optimal rate of decrease for $\lambda_n$, 
Thm. 1 of \citet{CAP07} yields the following convergence statements for the estimated embeddings, under assumptions to be discussed in Section~\ref{Assumptions}.
\begin{corollary} \label{Cor:CapUpper}
 Let $b = \infty, c> 1$ then for every $\epsilon >0$ there exists a constant $\tau$ such that
$$ 
 \limsup_{n \rightarrow \infty} \sup_{P \in \mathscr{P}(b,c)} P^n \left[\Err_s[\hat \mu_n] - \Err_s[\mu'] > \tau \frac{1}{n}  \right] < \epsilon.$$
Let $b=2$ and $c =1$ then for every $\epsilon >0$ there exists a constant $\tau$ such that
\begin{align*}
&
\limsup_{n \rightarrow \infty} \sup_{P \in \mathscr{P}(b,c)} P^n \left[\Err_s[\hat \mu_n] - \Err_s[\mu'] > \tau \left(\frac{\log n}{n} \right)^{\frac{2}{3}}  \right] \\ 
&< \epsilon.
\end{align*}
\end{corollary}
The rate for the estimate $\hat \mu_n$ can be complemented with minimax lower rates for vector valued regression \cite{CAP07}[Th. 2] in the case that $b < \infty$.
\begin{corollary}  \label{Cor:CapLower} 
Let $b=2$ and $c =1$ and let $\Lambda_n:= \{l_n \, |l_n: (\cX \times \cY)^n \rightarrow \cH_\G \}$ be the set of all learning algorithm working on $n$ samples, outputting $\nu_n\in\cH_\G$. Then for every $\epsilon >0$ there exists a constant $\tau > 0$ such that
\begin{align*}
&
\liminf_{n \rightarrow \infty} \inf_{l_n \in \Lambda_n} \sup_{P \in \mathscr{P}(b,c)} P^n \left[\Err_s[\nu_n] - \Err_s[\mu'] > \tau \left(\frac{1}{n} \right)^{\frac{2}{3}}  \right] \\ 
&> 1- \epsilon.
\end{align*}
\end{corollary}
This corollary tells us that there exists no learning algorithm which can achieve  better rates than $n^{-\frac{2}{3}}$ uniformly over $\mathscr P(2,1)$, and hence the estimate $\hat \mu_n$ is optimal up to a logarithmic factor.

\paragraph{State of the art results for the embedding}
The current convergence result for the embedding is proved by \citet[][Th.1]{DBLP:journals/jmlr/SongGG10}. A crucial assumption that we discuss in detail below is that the mapping $x \mapsto \Expect[h(Y) | X=x]$ is in the RKHS $\cH_K$ of the real valued kernel, i.e. that for all $h \in \cH_L$ we have that there exists a $f_h \in \cH_K$, such that 
\begin{equation} \label{eq:SongAssumption}
\Expect[h(Y)|X=x] =  f_h(x).
\end{equation}
The result of Song et al. implies the following  (see App.  \ref{sec:proofs}): if $K(x,x) < B$ for all $x \in \cX$ and the schedule $\lambda(n) = n^{-1/4}$ is used: 
for a fixed probability measure $P$, 
 there exists a constant $\tau$ such that
\begin{align} \label{ArthurConv}
 &\lim_{n \rightarrow \infty} P^n\left[ \Err[\hat \mu_n]  > \tau \left(\frac{1}{n}\right)^{\frac{1}{4}} \right]  = 0,
\end{align}
where $\hat \mu_n(x)$ is the estimate from Song et al. No complementary lower bounds were known until now.

\paragraph{Comparison}
The first thing to note is that under the assumption that  $\Expect[h | X]$ is in the RKHS $\cH_K$ the minimiser of $\Err_s$ and $\Err$ are a.e. equivalent due to Theorem \ref{thm:MinEqual}: the assumption implies  a $\mu^* \in \cH_\Gamma$ exists with $\Expect[h | X] = \langle h,\mu^*(X)\rangle_L$ for all $h \in \cV$ (see App. \ref{app:SongAss} for details). 
Hence, under this assumption, the statements from eq. \ref{ArthurConv} and  Cor. \ref{Cor:CapUpper} ensure we converge to the 
true conditional expectation, and achieve an error of 0 in the risk $\Err$.

In the case that this assumption is not fullfilled and eq. \ref{ArthurConv} is not applicable, Cor. \ref{Cor:CapUpper} still tells us that we converge to the minimiser of $\Err_s$. Coupling this statement with Thm. \ref{thm:bound} allows us
to bound the distance to the minimal error $\Err[\mu^*]$, where $ \mu^* \in \cH_\G$  minimises $\Err$.

The other main differences are obviously the rates, and
that Cor. \ref{Cor:CapUpper}  bounds the error
uniformly over a space of probability measures, while eq. \ref{ArthurConv} provides only a point-wise statement (i.e., for a fixed probability measure $P$). 

\subsection{Assumptions} \label{Assumptions}

\paragraph{Cor. \ref{Cor:CapUpper} and \ref{Cor:CapLower}} 
Our main assumption is that $\cH_L$ is finite dimensional. It is likely that this assumption can be weakened, but this requires a deeper analysis. 

The assumptions of  \citet{CAP07} are summarized in Table
\ref{tab:Ass}, where we provide  details in App. \ref{Ass:CapVerif}. App. \ref{sec:OurAss} contains simple and complete assumptions that ensure all statements in the table hold. Beside some measure theoretic issues, the assumptions 
are fulfilled if for example, 1) $\cX$ is a compact subset of $\R^n$, $\cY$ is compact, $\cH_L$ is a finite dimensional RKHS, $\G$ and $L$ are continuous; 2) $\mathcal{E}_s[\mu'] = \inf_{\mu \in \cH_\G} \mathcal{E}_s[\mu]$. This last condition is unintuitive, but can be rewritten in the following way: 
\begin{theorem}[Proof in App.]
Let  $||h||_\cV,||\mu(x) - h||_{\cV}$ be integrable for all $h \in \cH_\G$ and let $\cV$ be finite dimensional. Then there exists a $\mu' \in \cH_\G$ with 
$\mathcal{E}_s[\mu'] = \inf_{\mu \in \cH_\G} \mathcal{E}_s[\mu]$ iff a $B> 0$ exists and a sequence $\{\mu_n\}_{n \in \mathbb{N}}$ with
$\mathcal{E}_s[\mu_n] \leq \inf_{\mu \in \cH_\G} \mathcal{E}_s[\mu] +1/n $ and $||\mu_n||_\G < B$.
\end{theorem}
The intuition is  that the condition is not fulfilled
if we need to make $\mu_n$ more and more complex (in the sense of a high RKHS norm) to optimize the risk. 

\paragraph{Definition and discussion of $\mathscr{P}(b,c)$}
The family of probability measures $\mathscr{P}(b,c)$ is characterized through spectral properties of the kernel function $\Gamma$.  
The assumptions correspond to assumptions on the eigenvalues in Mercer's theorem in the real valued case, i.e. that there are finitely many eigenvalues or that the eigenvalues decrease with a certain rate. In detail, define the operator $A$ through $A(\phi)(x') := \int_\cX \Gamma(x',x) \phi(x) dP_\cX$, where $\phi \in L^2(P_\cX)$. $A$ can be written as \citep[Rem. 2]{CAP07}
$ A = \sum_{n=1}^N \lambda_n \langle \cdot,\phi_n \rangle_P \phi_n,$
where the inner product is the $L^2$ inner product with measure $P_\cX$ and $N=\infty$ is allowed. As in the real valued case, the eigendecomposition depends 
on the measure on the space $\cX$ but is independent of the distribution on $\cY$. The eigendecomposition measures the complexity of the kernel, where the lowest 
complexity is achieved for finite $N$ --- that is, the case $b=\infty,c>1$ --- and has highest complexity if the eigenvalues decrease with the slowest possible rate, $\lambda_n < C/n$ for a constant $C$. The case $b=2,c=1$ correspond to a slightly faster decay, namely, $\lambda_n < C/n^2$. In essence, there are no assumptions 
on the distribution on $\cY$, but only on the complexity of  the kernel $\Gamma$ as measured with $P_\cX$.



\paragraph{Embedding}
The results of \citet{DBLP:journals/jmlr/SongGG10} do not rely on the assumption 
that $\cV$ is finite dimensional.
Other conditions on the distribution are required, however, which are challenging to verify.
To describe these conditions, we recall the real-valued RKHS $\cH_K$ with kernel $K$, and define the uncentred cross-covariance operator $C_{YX}:\:\cH_K\rightarrow\cH_L$ such that $\langle g, C_{YX} f\rangle_{\cH_L}=\Expect_{XY}(f(X)g(Y))$, with the covariance operator $C_{XX}$ defined by analogy.
One of the two main assumptions of Song et al. is that $C_{YX} C_{XX}^{-3/2}$ needs to be Hilbert-Schmidt. The covariances $C_{YX}$ and $C_{XX}$ are compact operators, meaning
$C_{XX}$ is not invertible when $\cH_K$ is  infinite dimensional
  (this gives rise to a notational issue, although the ``product'' operator $C_{YX} C_{XX}^{-3/2}$ may still be defined). Whether $C_{YX} C_{XX}^{-3/2}$ is Hilbert-Schmidt (or even bounded) will depend on the underlying distribution $P_{XY}$ and on the kernels $K$ and $L$.
At this point, however, there is no easy way to translate properties of $P_{XY}$ to guarantees that the assumption holds.

The  second main assumption is that the conditional expectation can be represented as an RKHS element (see App \ref{app:SongAss}). 
Even for rich RKHSs (such as universal RKHSs), 
it can be challenging to determine the associated conditions on the distribution $P_{XY}$.
For simple finite dimensional RKHSs, the  assumption may fail, as shown below.
\begin{corollary}[Proof in App. \ref{sec:proofs} ]
Let $\cV$ be finite dimensional such that a function $\tilde h \in \cV$ exists
with $\tilde h(y) \geq \epsilon > 0$ for all $y \in \cY$. Furthermore, let
$\cX := [-1,1]$ and the reproducing kernel for $\cH_K$ be $K(x,y) = xy$. Then there
exists no measure for which the assumption from eq. (\ref{eq:SongAssumption}) can be fulfilled. 
\end{corollary}

\section{Sparse embeddings}
In many practical situations it is desirable to approximate the conditional mean embedding by a sparse version which involves a smaller number of parameters. For example, in the context of reinforcement learning and planning, the sample size $n$ is large and we want to use the embeddings over and over again, possibly on many different tasks and over a long time frame.

Here we present a technique to achieve a sparse approximation of the sample mean embedding. Recall that this 
is given by the formula (cf. equation \eqref{MuForm}) 
$$
{\hat \mu}(x) = \sum_{i,j=1}^n W_{ij} K(x_i,x) L(y_j,\cdot),
$$
where $W = (K+n\lambda I)^{-1}$. A natural approach to find a sparse approximation of ${\hat \mu}$ is to look for a function $\mu$ which is close to ${\hat \mu}$ according to the RKHS norm $\|\cdot\|_\G$
(in App. \ref{app:massi} we establish a link between this objective and our cost function $\Err$). In the special case that $\G = K Id$ this amounts to solving the optimization problem
\begin{equation}
\min_{M \in \R^{n\times n}} 
f(M-W) + \gamma \|M\|_{1,1} 
\label{eq:lollo}
\end{equation}
where $\gamma$ is a positive parameter, $\|W\|_{1,1} :=\sum_{i,j}^n |M_{ij}|$ and
\begin{equation}
f(M)=\Big\|\sum_{i,j= 1}^n M_{ij} K(x_i,\cdot) L(y_j,\cdot)\Big\|_{K\otimes L} \; .
\label{eq:exp_norm}
\end{equation}
Problem \eqref{eq:lollo} is equivalent to a kind of Lasso problem with $n^2$ variables: when $\gamma = 0$, $M=W$ at the optimum and the approximation error is zero, however as $\gamma$ increases, the approximation error increases as well, but the solution obtained becomes sparse (many of the elements of matrix $M$ are equal to zero).

A direct computation yields that the above optimization problem is equivalent to
\begin{equation}
\min_{M \in \R^{n\times n}} {\rm tr} ((M-W)^\trans K (M-W) L) + \gamma \sum_{i,j=1}^n |M_{ij}|.
\label{eq:Mlasso}
\end{equation}
In the experiments in Section \ref{sec:exp}, we solve problem \eqref{eq:Mlasso} with FISTA \citep{fista}, an 
optimal first order method which requires $O(1/\sqrt{\epsilon})$ iterations to reach a $\epsilon$ accuracy 
of the minimum value in \eqref{eq:Mlasso}, with a cost per iteration of $O(n^2)$ in our case. The algorithm is outlined below, where $S_\gamma(Z_{ij}) = \text{sign}(Z_{ij})(|Z_{ij}| - \gamma)_+$ and $(z)_+ = z$ if $z > 0$ and zero otherwise.

\begin{algorithm}
\caption{LASSO-like Algorithm}
\label{lasso_algo}
\begin{algorithmic}
\STATE {\bf input:} $W$, $\gamma$, $K$, $L$ {\bf output:} $M$
\STATE $Z_1 = Q_1 = 0, \theta_1 = 1, C = \|K\|~\|L\|$
\FOR {t=1,2,\dots}
\STATE $Z_{t+1} = S_{\gamma C}\left(Q_{t} - C~(K~Q_{t}~L - G~W~L)\right)$
\STATE $\theta_{t+1} = \frac{1+\sqrt{1+4\theta_t^2}}{2}$
\STATE $Q_{t+1} = Z_{t+1} + \frac{\theta_t - 1}{\theta_{t+1}}(Z_t - Z_{t+1})$
\ENDFOR 
\end{algorithmic}
\end{algorithm}

Other sparsity methods could also be employed. For example, we may replace the norm $\|\cdot\|_{1,1}$ by a block $\ell_1/\ell_2$ norm. That is, we may choose the 
norm $\|M\|_{2,1} :=\sum_{j=1}^n \sqrt{\sum_{i=1}^n M_{ij}^2}$, which is the sum of the $\ell_2$ norms of the rows of $M$. 
This penalty function  encourages sparse approximations which use few input points but all the outputs. 
Similarly, the penalty $\|M^\trans\|_{2,1}$ will sparsify over the outputs. Finaly, if we wish to remove many pair of examples we may use the more sophisticated penalty $\sum_{i,j=1}^n \sqrt{\sum_{k=1}^m M_{ik}^2 + M_{kj}^2
}$.

\vspace{-0.3cm}
\begin{figure}[t]
\centering
\includegraphics[scale=0.4]{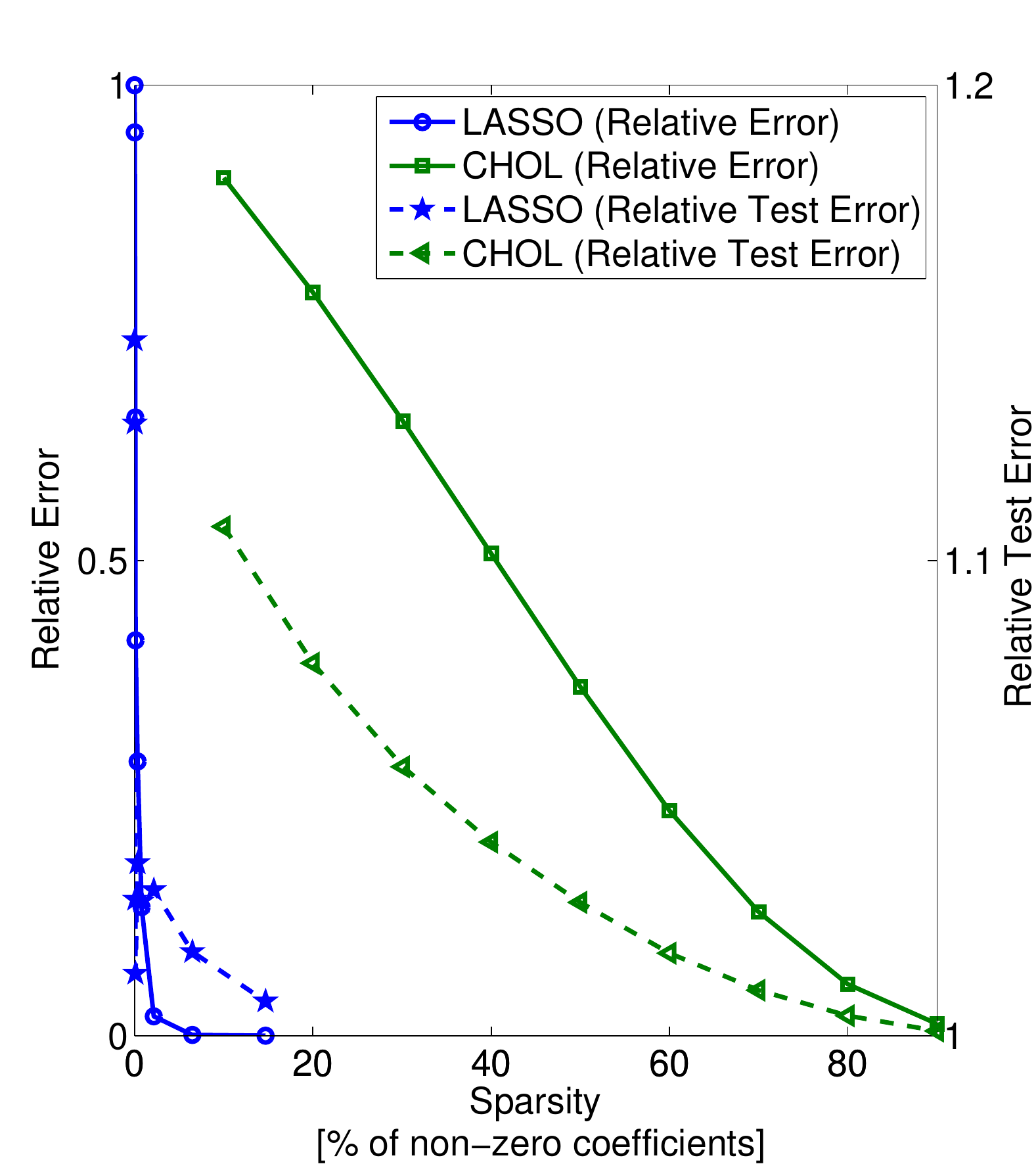}
\vspace{-3mm}
\caption{Comparison between our sparse algorithm and an incomplete Cholesky decomposition. The x-axis shows the level of sparsity, where on the right side the original 
solution is recovered. The y-axis shows the distance to the dense solution and the test error relative to the dense solution.  } \label{fig:comparison}
\end{figure}

\subsection{Experiment} \label{sec:exp}
We demonstrate here that the link between vector-valued regression and
the mean embeddings can be leveraged to develop useful embedding alternatives that exploit properties of the regression formulation: we apply the sparse algorithm to a challenging  reinforcement learning task. The sparse algorithm 
makes use of the labels, while other algorithms to sparsify the embeddings without our regression interpretation cannot make use of these. In particular, a popular method to sparsify is the incomplete Cholesky decomposition \cite{shawetaylor2004kernel}, which sparsifies based on the distribution on the input space $\cX$ only. We compare to this method in the experiments.

The reinforcement learning task is the under-actuated pendulum swing-up problem
from \citet{DBLP:journals/ijon/DeisenrothRP09}. We generate a discrete-time approximation of the continuous-time pendulum dynamics as done in \cite{DBLP:journals/ijon/DeisenrothRP09}. Starting from an arbitrary state the goal is to swing the pendulum up and balance it in the inverted position. The applied torque is $u \in [-5,5]Nm$ and is not sufficient for a direct swing up. The state space is defined by the angle $\theta \in [-\pi, \pi]$ and the angular velocity, $\omega \in [-7,7]$. The reward is given by the function $r(\theta,\omega) = \mathrm{exp}(-\theta^2 - 0.2\omega^2)$.
The learning algorithm is a kernel method which uses the mean embedding estimator to perform policy iteration \cite{RLEmbNoStar}. Sparse solutions are in this task very useful as the policy iteration 
applies the mean embedding many times to perform updates. 
The input space has $4$ dimensions (sine and cosine of the angle, angular velocity and applied torque), while the output has $3$ (sine and cosine of the angle and angular velocity). We sample uniformly a training set of $200$ examples and learn the mean conditional embedding using the direct method \eqref{EmbeddingExpansion}. We then compare the sparse approximation obtained by Algorithm \ref{lasso_algo} using different values of the parameter $\gamma$ to the approximation obtained via an incomplete Cholesky decomposition at different levels of sparsity. We assess the approximations using the test error and \eqref{eq:exp_norm}, which is an upper bound on the generalization error (see App. \ref{app:massi}) and report the results in Figure \ref{fig:comparison}.

\section{Outlook}
We have established a link between vector-valued regression and conditional mean embeddings. 
On the basis of this link,
 we derived a sparse embedding algorithm, showed how cross-validation can be performed,  established better convergence rates under milder assumptions, and complemented these upper rates with lower rates, showing that the embedding estimator achieves  near optimal rates.


There are a number of interesting questions and problems which follow from
 our framework. It may be valuable to employ other
kernels $\G$ in place of the kernel $K(x,y) \Id$ that leads to the mean embedding, so as to exploit knowledge about the data generating process. As a related observation, for the kernel $\G(x,y) := K(x,y) \Id$, $\G_x$ is not 
a Hilbert-Schmidt operator if $\cV$ is infinite dimensional, as
\begin{align*}
&||\G_x||^2_{HS} = K(x,x) \sum_{i=1}^\infty \langle e_i, \Id e_i \rangle_L = \infty,
\end{align*}
however the convergence results from \cite{CAP07} assume $\G_x$ to be Hilbert-Schmidt. 
While this might simply be a result of
the  technique used in \cite{CAP07}, it might also indicate a deeper problem with the standard embedding estimator, namely that if $\cV$ is infinite dimensional then
 the rates degrade. The latter case would have a natural interpretation as an over-fitting effect, as $\Id$ does not ``smooth'' the element $h \in \cV$. 


Our sparsity approach can potentially be equipped with other regularisers that cut down  on rows and columns of the $W$ matrix in parallel.  Certainly, ours is not the only 
sparse  regression approach, and other sparse regularizers might yield good performance on appropriate problems.

\section*{Acknowledgements}
The authors want to thank for the support of the EPSRC \#EP/H017402/1 (CARDyAL) and the European Union \#FP7-ICT-270327 (Complacs).

\begin{small}
\bibliographystyle{icml2012}
\bibliography{reinforcement,guygraph,PoliciesBib}

\end{small}

\begin{onecolumn}
\begin{center}
\begin{huge}
\underline{\textbf{SUPPLEMENTARY}}
\end{huge}
\end{center}
\appendix

\section{Similarity of minimisers}
\label{sec:MinEqual}
%
We assume that all $h \in \cV$ are integrable with respect to the regular conditional probability $P(\cdot|x)$ and all $\mu \in \cH_\G$ are  integrable with respect
to $\tilde P$. In particular, these are fulfilled if our 
general assumptions from Section \ref{sec:OurAss} are fulfilled.
\begin{lemma} \label{lem:MinHelper}
If there exists  $\mu^* \in \cH_\G$ such that for any $h\in \cV$: 
$\Expect[h |X] = \lang h,  \mu^*(X) \rang_\cV$ $P_\cX$-a.s., then for any
$\mu \in \cH_\G$:
\begin{align*}
&(i) \quad \Expect_{X,Y} \langle L(Y,\cdot), \mu^*(X) \rangle_\cV = \Expect_{X} || \mu^*(X)||^2_\cV,  \\
&(ii) \quad \Expect_{X,Y} \langle L(Y,\cdot), \mu(X) \rangle_\cV = \Expect_{X} \langle \mu^*(X), \mu(X) \rangle_\cV.
\end{align*}
\end{lemma}
\begin{proof}
(i) follows from (ii) by setting $\mu := \mu^*$. (ii) can be derived like that:
\begin{align*}
&
\Expect_{X} \langle \mu(X), \mu^*(X)\rangle_\cV =
 \Expect_{X} \Expect_Y[\mu(X)(Y)| X] = \Expect_{X,Y} \langle L(Y,\cdot), \mu(X)\rangle_\cV, 
\end{align*}
where we used Fubini's theorem for the last equality \citep[Thm. 1.27]{KALL01}.
\end{proof}

\begin{theorem}[Thm  \ref{thm:MinEqual}]
If there exists a $\mu^* \in \cH_\G$ such that for any $h\in \cV$: 
$\Expect[h |X] = \lang h,  \mu^*(X) \rang_\cV$ $P_\cX$-a.s.,
then 
\begin{align*}
&\mu^* = \argmin_{\mu \in \cH_\G} \Err[\mu]
= \argmin_{\mu \in \cH_\G} \Err_s[\mu]
\quad P_\cX \, a.s..
\end{align*}
Furthermore, the minimiser is $P_\cX$-a.s. unique.
\end{theorem}
\begin{proof}
We start by showing that the right side is minimised by $\mu^*$. Let $\mu$ be any element in $\cH_\G$ then we have
\begin{align*}
&\Expect_{X,Y}\left[ || L(Y,\cdot) - \mu(X)||_\cV^2  \right] 
-  \Expect_{X,Y}\left[ || L(Y,\cdot) - \mu^*(X)||_\cV^2  \right]\\
&= \Expect_{X} || \mu(X)||_\cV^2  
-2 \Expect_{X,Y} \langle L(Y,\cdot), \mu(X)\rangle_\cV  
+ 2\Expect_{X,Y} \langle L(Y,\cdot), \mu^*(X)\rangle_\cV 
-  \Expect_{X} || \mu^*(X)||_\cV^2 \\
&= \Expect_{X} || \mu(X)||_\cV^2  
-2 \Expect_{X} \langle \mu^*(X), \mu(X)\rangle_\cV  
+  \Expect_{X} || \mu^*(X)||_\cV^2 
= \Expect_{X}\left[ || \mu(X) - \mu^*(X)||_\cV^2 \right]  \geq 0.
\end{align*}
Hence, $\mu^*$ is a minimiser. The minimiser is furthermore $P_\cX$-a.s. unique: Assume there is a second minimiser $\mu'$ then above calculation shows that $ 0 = \Expect_{X,Y}\left[ || L(Y,\cdot) - \mu'(X)||_\cV^2  \right]  -  \Expect_{X,Y}\left[ || L(Y,\cdot) - \mu^*(X)||_\cV^2  \right] =  \Expect_{X}\left[ || \mu^*(X) - \mu'(X)||_\cV^2 \right] =  \Expect_{X}\left[ || \mu^*(X) - \mu'(X)||_\cV^2 \right]$. Hence,  $|| \mu^*(X) - \mu'(X)||_\cV = 0$  $P_\cX$-a.s. \cite{FREM00}[122Rc], i.e. a measurable 
set $M$ with $P_\cX M = 1$ exists such that $|| \mu^*(X) - \mu'(X)||_\cV = 0$ holds for all $X\in M$. As $||\cdot ||_\cV$ is 
a norm we have that $\mu^*(X) = \mu'(X)$ $P_\cX$-a.s..

To show the second equality
we note that, for every $h \in \cV$, 
$ \Expect_{X}(\Expect[h |X] - \lang h , \mu^*(X) \rang_\cV )^2 = 0 $ by assumption. Hence, the supremum over $\cB(\cV)$ is also 0 and the minimum is attained at $\mu^*(X)$. Uniqueness can be seen in the following way:
assume there is a second minimiser $\mu'$ then for all $h \in \cH_\cV$  we have 
$\Expect_X (\lang h , \mu'(X) - \mu^*(X) \rang_\cV  )^2 \leq
\Expect_{X}(\lang h , \mu'(X) \rang_\cV - \Expect[h|X])^2 +  
\Expect_{X}(\Expect[h |X] - \lang h , \mu^*(X) \rang_\cV )^2 = 0$. 
Hence, $\lang h , \mu'(X) - \mu^*(X) \rang_\cV = 0$  $P_\cX$-a.s. \cite{FREM00}[122Rc], i.e. a measurable 
set $M$ with $P_\cX M = 1$ exists such that $\lang h , \mu'(X) - \mu^*(X) \rang_\cV = 0$ holds for all $X\in M$. Assume 
that there exists a $X' \in M$ such that $\mu'(X') \not = \mu^*(X')$ then pick $h:= \mu'(X') - \mu^*(X')$
and we have  $0 = \lang h , \mu'(X') - \mu^*(X') \rang_\cV = ||\mu'(X') - \mu^*(X') ||_\cV > 0$ as 
$||\cdot||_\cV$ is a norm. Hence, $\mu'(X') = \mu^*(X')$ a.s..
\end{proof}

\begin{theorem}
If there exists $\eta>0$ and  $\mu^* \in \cH_\G$ such that  
$\sup_{\mu \in \cB(\cH_\G)}\Expect_X \left[ \Expect[ \mu(X) |X]  -  \lang \mu(X) ,  \mu^*(X) \rang_\cV\right]^2 = \eta < \infty$, that  $\mu'$ is a minimiser  of $\Err_s$  and $\hat \mu$ is an element of $\cH_\G$ with 
$\Err_s[\hat \mu] \leq \Err_s[\mu'] + \delta$ then 
\begin{align*}
&(i) \quad \Err[\mu']  \leq 
\left(\sqrt{\Err[\mu^*]} + \eta^{1/4}\sqrt{8 ( \|\mu^*\|_\G 
+\|\mu'\|_\G)}\right)^2 \\
&(ii) \quad 
\Err[\hat \mu]  \leq 
\left(\sqrt{\Err[\mu^*]} + \eta^{1/4}\sqrt{8 ( \|\mu^*\|_\G 
+\|\hat \mu\|_\G)} + \delta^{1/2}\right)^2.
\end{align*}
\end{theorem} 
\begin{proof}
First, observe that if $\mu \in \cH_\G$ then we have due to the Jensen inequality
\begin{align*}
&|\Expect_X \Expect[ \mu(X) |X]  -  \Expect_X \langle \mu(X) ,  \mu^*(X) \rangle_\cV| 
\leq ||\mu||_\G \Expect_X \left| \Expect\left[ \frac{\mu(X)}{||\mu||_\G }|X\right]  -  \langle \frac{\mu(X)}{\|\mu\|_\G} ,  \mu^*(X) \rangle_\cV \right| \\
&\leq ||\mu||_\G \sqrt{ \Expect_X \left( \Expect\left[ \frac{\mu(X)}{||\mu||_\G }|X\right]  -  \langle \frac{\mu(X)}{\|\mu\|_\G} ,  \mu^*(X) \rangle_\cV\right)^2 } 
= ||\mu||_\G \sqrt{\eta}.
\end{align*}
We can now reproduce the proof of Lemma \ref{lem:MinHelper} with an approximation error. For any $\mu \in \cH_\G$ we have
\begin{align*}
&|\Expect_{X} \langle \mu(X), \mu^*(X)\rangle_\cV - \Expect_{X,Y} \langle L(Y,\cdot), \mu(X)\rangle_\cV| =  |\Expect_{X} \langle \mu(X), \mu^*(X)\rangle_\cV - \Expect_X \Expect[\mu(X)| X] | \leq \|\mu\|_\G \sqrt{\eta}.  
\end{align*}
In particular, 
\begin{align*}
&|\Expect_{X,Y} \langle L(y,\cdot), \mu^*(X) \rangle_\cV  -  \Expect_{X,Y} || \mu^*(X)||^2_\cV | \leq \|\mu^*\|_\G \sqrt{\eta}.  
\end{align*}
Like in the proof of Thm. \ref{thm:MinEqual} we have for any $\mu$ that  
\begin{align}
& \Err_s[\mu] - \Err_s[\mu^*] = 
\Expect_{X,Y}\left[ || L(Y,\cdot) - \mu(X)||_\cV^2  \right] 
-  \Expect_{X,Y}\left[ || L(Y,\cdot) - \mu^*(X)||_\cV^2  \right]  \nn \\
&\geq \Expect_{X} || \mu(X)||_\cV^2  
-2 \Expect_{X} \langle \mu^*(X), \mu(X)\rangle_\cV  
+  \Expect_{X} || \mu^*(X)||_\cV^2 - 2 \|\mu^*\|_\G \sqrt{\eta}
-2 \|\mu\|_\G \sqrt{\eta}  \nn \\
&= \Expect_{X}\left[ || \mu^*(X) - \mu(X)||_\cV^2 \right]  
- 2\sqrt{\eta}( \|\mu^*\|_\G 
+\|\mu\|_\G ). \label{eq:CostRelOneDir}
\end{align}
In particular, 
$| \Err_s[\mu] - \Err_s[\mu^*]| \geq   \Err_s[\mu] - \Err_s[\mu^*] \geq
 \Expect_{X}\left[ || \mu^*(X) - \mu(X)||_\cV^2 \right]  
- 2\sqrt{\eta}( \|\mu^*\|_\G 
+\|\mu\|_\G )$ and hence 
\begin{equation} 
\Expect_{X}\left[ || \mu^*(X) - \mu(X)||_\cV^2 \right] \leq
| \Err_s[\mu] - \Err_s[\mu^*]| + 2\sqrt{\eta}( \|\mu^*\|_\G 
+\|\mu\|_\G ). \label{eq:CostSim}  
\end{equation}
We can now bound the error $\Err[\mu]$ in dependence of how similar 
$\mu$ is to $\mu^*$ in the surrogate cost function $\Err_s$:
\begin{align}
&\sqrt{\Err[\mu]} = \sqrt{\Err[\mu^*]} + \sup_{h \in \cB(\cV)} \sqrt{\Expect_X \left[
 \lang h  ,  \mu^*(X) - \mu(X) \rang_\cV \right]^2}
\leq \sqrt{\Err[\mu^*]} +  \sqrt{\Expect_X \left(\frac{|| \mu^*(X) -  \mu(X) ||^2_\cV}{||\mu^*(X) -  \mu(X)||_\cV}  \right)^2} \nn \\
&\leq \sqrt{\Err[\mu^*]} + \eta^{1/4}\sqrt{2 ( \|\mu^*\|_\G 
+\|\mu\|_\G)} + 
\sqrt{ | \Err_s[\mu] - \Err_s[\mu^*] | 
}, \label{eq:FinalEqForCost}
\end{align}
where we used that $ \lang  \frac{\mu^*(X) -  \mu(X)}{||\mu^*(X) -  \mu(X) ||_\cV },  \mu^*(X) -  \mu(X) \rang_\cV \geq 
\lang  h,  \mu^*(X) -  \mu(X) \rang_\cV$ for any $h \in \cB(\cV)$ (observe that $h$ is independent of $X$) and eq. \ref{eq:CostSim}.

Now, for $\mu := \mu'$ observe that  $ \Err_s[\mu']  + 2\sqrt{\eta}( \|\mu^*\|_\G 
+\|\mu\|_\G )
\geq \Err_s[\mu^*] $ follows from eq. (\ref{eq:CostRelOneDir})
 and as $\mu'$ is a $\Err_s$ minmiser we have 
$  | \, \Err[\mu^*] - \Err[\mu']  \, |  \leq  2\sqrt{\eta}( \|\mu^*\|_\G 
+\|\mu\|_\G )$ and from eq. \ref{eq:FinalEqForCost} we get
\begin{align*}
 &
 \sqrt{\Err[\mu']} \leq \sqrt{\Err[\mu^*]} + \eta^{1/4}\sqrt{8 ( \|\mu^*\|_\G 
+\|\mu'\|_\G)}. 
\end{align*}
Similarly,  for $\mu := \hat \mu$ we have
\begin{align*}
& \sqrt{\Err[\hat \mu]} \leq \sqrt{\Err[\mu^*]} + \eta^{1/4}\sqrt{8 ( \|\mu^*\|_\G + \|\hat \mu\|_\G)} + \delta^{1/2}.
\end{align*}
\end{proof}

\section{Assumptions}
\subsection{Simple assumptions to make the convergence theorems hold.} \label{sec:OurAss}



In this section, we present general assumptions that make the assumptions in \citep[Theorems 1 and 2]{CAP07} hold.  Weaker assumptions are certainly possible  and we do not claim that our set of assumptions are the most general possible.

\subsubsection{Assumptions for the involved spaces and kernels}
We assume that $\cX$ is a compact subset of $\R^n$, $\cY$ is a compact set and $\cV$ is a finite dimensional RKHS. Furthermore, we assume that we work on measure spaces $(\R,\mathscr{B}(\R)),(\cX,\mathscr{B}(\cX)), (\cY,\mathscr{B}(\cY)),
(\cV,\mathscr{B}(\cV))$, where $\mathscr{B}(\R)$ etc. are the respective Borel algebras and each of the spaces
is equipped with its norm induced topology.  
Finally, we assume that $\G$ and $L$ are continuous. Hence, $L(y,y) \leq B < \infty$ for all $y \in \cY$.

\subsubsection{Data generating distribution} \label{sec:DataGen}
The main object in this study is the probability measure $\tilde P$ on the space $\cX \times \cY$ from which the data is generated. 
Formally, we have a measure space\footnote{Some measure theory is needed to write things down cleanly. See also \cite{Steinwart:2008:SVM:1481236}[chp. 2] where similar problems are discussed.}
 $(\cX\times \cY, \tilde \Sigma, \tilde P)$ where $\tilde \Sigma$ is a suitable $\sigma$-algebra and
$\tilde P:\tilde \Sigma \mapsto [0,1]$ a probability measure. Now we have a transformation from $\phi: y \mapsto L(y,\cdot)$ which turns a sample $\{(x_1,y_1), (x_2, y_2), \ldots \}$
into 
$\{(x_1,L(y_1,\cdot)), (x_2, L(y_2,\cdot), \ldots \}$ with $(x_i,L(y_i,\cdot)) \in \cX \times \cV$. The measure $\tilde P$ and the $\sigma$-algebra
$\tilde \Sigma$ induce through the image measure construction a measure space $(\cX \times \cV,\Sigma,P)$, where $\Sigma := \{E | \phi^{-1}[E] \in \tilde \Sigma, E \subset \cX \times \cV\}$ 
is the largest $\sigma$-algebra on $\cX \times \cV$ such that $\phi$ is measurable and $P E = \tilde P(\phi^{-1}[E])$ for all $E \in \Sigma$ \cite{FREM00}[112E].
 
We assumed already that $\cX,\cV$ carry the Borel algebra. In addition  assume
that $\tilde \Sigma = \mathscr{B}(\cX \times \cY)$ and 
 $\Sigma = \mathscr{B}(\cX \times \cV)$, i.e. the Borel algebra on the product spaces. 
The assumptions we stated till now guarantee us that the regular conditional  probabilities $P(B|x)$ and $\tilde P(\tilde B|x)$ exists,
where $B \in \mathscr{B}(\cV)$ and $\tilde B \in \mathscr{B}(\cY)$ 
 \cite{Steinwart:2008:SVM:1481236}[Lem. A.3.16]. 

Our assumptions on $L$ and $\G$ guarantee that all $h$ in $\cV$  and $\mu \in \cH_\G$ are continuous \cite{BER04}[Th. 17] and Cor. \ref{Cor:ContRKHS}. Hence, all $h$ and $\mu$ are measurable as by assumption all spaces are equipped with Borel algebras \cite{FREM03}[4A3Cb,d]. Furthermore, as each $|h|$ and $|\mu|$ is a continuous function on a compact set it is upper bounded by a constant $B$. Thus all $h$ and $\mu$ are integrable with respect  to any probability measure on $(\cY,\mathscr{B}(\cY))$ and $(\cX \times \cY,\mathscr{B}(\cX \times \cY))$.
 
The final assumption is that one of the two equivalent conditions
of theorem \ref{thm:infAttained} is fulfilled.



\subsection{Verification of the Caponnetto \& De Vito assumptions} \label{Ass:CapVerif}
Theorem 1-2 from \cite{CAP07} are based on a set of assumptions. 
The assumptions can be grouped into three categories:
first, assumptions about the space $\cX$ and $\cV$, second assumptions about the space of regression functions that we use and third about the underlying true probability distribution.
Table \ref{tab:Ass} summarizes the assumptions that need to be fulfilled in our setting.
In the following we discuss each assumption shortly and we show ways to verify them. 
Also note, that the general assumption from the last section guarantee that all assumptions in table \ref{tab:Ass} are fulfilled.

\paragraph{Assumptions for $\cX$ and $\cV$}
$\cX$ needs to be a Polish space, that is a separable completely metrizable topological space.
If $\cX$ is, for example, a subset of $\R^n$ equipped with any metric then $\cX$ is Polish if $\cX$ is closed.   

$\cV$ needs to be a separable Hilbert space to be able
to apply the theorems. In our case $\cV$ is an RKHS so certainly a Hilbert space. Not all RKHSs are separable, but, due to our assumption
that $\cV$ is finite dimensional it follows that it is separable (e.g. take a basis $e_1, \ldots, e_n$ of $\cV$ then the countable set $\{\sum_{i\leq n} q_i e_i | q_i \in \mathbb{Q}  \}$ is dense in the RKHS norm).

Finally, $\text{Tr}(\G(x,x)) \leq B$ needs to hold for all $x \in \cX$, i.e. the trace needs to be bounded over $\cX$. 

\paragraph{Assumptions for $\cH_\G$}
$\cH_\G$ needs to be separable. Separability of $\cH_\G$ and continuity of the kernel are closely related in the scalar case \cite{BER04}[Sec. 1.5]. Similarly, we have: 
\begin{theorem} \label{Thm:HKSep}
$\cH_\G$ is separable if $\cX$ is a Polish space, $\cV$ is a finite dimensional Hilbert space and $\G$ is continuous in each argument with respect to the topology on $\cX$.
\end{theorem}

\begin{corollary}
If $\cX = \mathbb{R}^n$, $\cV$ is a finite dimensional RKHS and $\G$ is a continuous
kernel then $\cH_\G$ is separable.
\end{corollary}

Furthermore, the point evaluator $\G_x^*(f) = f(x)$, $ f \in \cH_\G$, $\G_x^*: \cH_\G \mapsto \cV$  needs to be Hilbert-Schmidt for every $x$. This is in our case always fulfilled as $\cV$ is finite dimensional: 
Let $\{e_i\}_{i \leq n}$ be an orthonormal basis of $\cV$:
\begin{align*}
&||\G_x^*||^2_{HS} = ||\G_x||^2_{HS} =  \sum_{i\leq n} ||\G_x e_i ||^2_\G =\sum_{i\leq n} \langle \G_x e_i, \G_x e_i \rangle_\G 
= \sum_{i \leq n} \langle e_i, \G(x,x) e_i \rangle_\cV. 
\end{align*}
But, $\G(x,x)$ is a finite dimensional matrix as $\cV$ is finite dimensional and hence the sum is finite and $\G_x$ has finite Hilbert-Schmidt norm.

The next condition concerns the measurebility of the kernel,  
$\forall f,g \in \cV, B:(x,x') \rightarrow  \langle  f, \G(x',x) g \rangle_\cV $ 
needs to be measurable. The simplest case where this holds is where $\G$ is continuous as a map from $(y,x)$ to $\mathcal{L}(\cV)$ and where we 
equip $\R, \cX$ and $\cV$ with the Borel algebra. Then we have measurebility as 
compositions from continuous functions are continuous and continuous functions are Borel measurable. 

\subsection{Assumptions for the ``true'' probability measure} \label{Sec:AssTrueProb}

There are further assumption concerning the measures $P$ and $\tilde P$ from which the data is generated that need 
to be fulfilled to be able to apply the convergence theorems. We now follow up on the discussion of these  measures in appendix \ref{sec:DataGen}. Again our general assumptions are sufficient 
to guarantee that the assumptions are fulfilled.

The first property that $P$ needs to fulfill is that
$ \int_{\cX \times \cV} ||h||_{\cV}^2 dP(x,h) < \infty $
where the integral is the Lebesgue integral.  This is essentially trivially fulfilled for all $P$ in our setting as $P$ is concentrated on elements $L(y,\cdot)$, i.e. 
$\cL = \{ L(y,\cdot)|y \in \cY\} \in \Sigma$ as $\phi^{-1}[\cL] = \cX \times \cY \in \tilde \Sigma$ and $P \cL = \tilde P(\phi^{-1} \cL) = \tilde P(\cX \times \cY) = 1$.
Furthermore, $||L(y,\cdot)||_{\cV} = L(y,y) < B < \infty$ if we have a bounded kernel $L$. The only problem is that we need to guarantee that $||h||_{\cV}^2$ is integrable. 
Let us therefore assume in the following 
that  $||h||_{\cV}^2 \upharpoonright \cL$ (i.e. the restriction to $\cL$) is measurable. We have $||h||_{\cV}^2 \upharpoonright \cL < B \chi \cL $
where $\chi$ is the characteristic function. $B \chi \cL$ is measurable as $\cL$ is measurable. Furthermore, it is integrable as it is a simple function and $P \cL = 1 < \infty$. Now, 
$||h\upharpoonright \cL||_{\cV}^2 $ is defined on the conegibile set $\cL$ (i.e. $P \cL = 1$) and by assumption $||h||_{\cV}^2 \upharpoonright \cL$ is measurable. Hence,
 $||h||_{\cV}^2 \upharpoonright \cL$ is integrable \cite{FREM00}[Lem. 122Jb] and as $||h||_{\cV}^2 \upharpoonright \cL =_{P - a.e.} ||h||_{\cV}^2$ we have that $||h||_{\cV}^2$ is integrable 
 \cite{FREM00}[122Ld] with integral $\int_{\cX \times \cV} ||h||_{\cV}^2 dP(x,v) \leq B < \infty$.

The next assumption concerns the infimum $ \mathcal{E}[\mu^*] = \inf_{\mu \in \cH_\G} \mathcal{E}[\mu],$
i.e. the infimum must be attained at a $\mu^* \in \cH_\G$. One can restate this condition in terms of the RKHS norm of a sequence converging to the 
infimum: the infimum is attained iff such a sequence exists which is bounded in the RKHS norm. The intuition is here that the condition is not fulfilled
if we need to make $\mu_n$ more and more complex (in the sense of a high RKHS norm) to optimize the risk. 

\begin{theorem} \label{thm:infAttained}
Let  $||h||_\cV^2,||\mu(x) - h||_{\cV}$ be integrable for all $\mu \in \cH_\G$ and let $\cV$ be a finite dimensional Hilbert space. Then there exists a $\mu^* \in \cH_\G$ with 
$\mathcal{E}[\mu^*] = \inf_{\mu \in \cH_\G} \mathcal{E}[\mu]$ iff a $B> 0$ exists and a sequence $\{\mu_n\}_{n \in \mathbb{N}}$ with
$\mathcal{E}[\mu_n] \leq \inf_{\mu \in \cH_\G} \mathcal{E}[\mu] +1/n $ and $||\mu_n||_\G < B$.
\end{theorem}
\begin{remark}
The important assumption is that the sequence $\{\mu_n \}$ is bounded in the RKHS norm as we can by definition always find a sequence $\{\mu_n\}$ that converges 
to the infimum. 
\end{remark}
\begin{proof}
``$\Rightarrow$'': set $\mu_n:=\mu^*$ then obviously the sequence converges to the infimum and is bounded as $|| \mu_n||_\G = ||\mu^*||_\G < \infty$ as $\mu^* \in \cH_\G$.

``$\Leftarrow$'':
We need to pull the limit through the integral and we need to make sure that the limit is attained by an RKHS function. First, let us check for the limit to be in the RKHS.
$\{\mu_n\}_{n\in \mathbb{N}}$ is a bounded sequence in a reflexive space (every Hilbert space is reflexive). Hence, there is a sub sequence $\{\mu_{n_k}\}_{k \in \mathbb{N}}$ which 
converges weakly  \cite{WER02}[Th. III.3.7]. Furthermore, $\cH_\G$ is closed and convex. Therefore, the weak limit $\mu$ is in $\cH_\G$ \cite{WER02}[Thm III.3.8].

Now we turn to the integral. We want to apply the Lebesgue dominated convergence theorem to move the limit inside the integral. We need to construct a suitable sequence: 
 Consider the sequence $s_k(x,h) := \{||g_k(x) - h||_\cV\}_{k\in \mathbb{N}}$ in $\R$, where $g_k := \mu_{n_k}$ and $\{\mu_{n_k}\}_{k \in \mathbb{N}}$ is the convergent sub sequence. 
The sequence  $\{s_k(x,v)\}_{k\in \mathbb{N}}$ is a Cauchy sequence in $\R$ as for any $\epsilon > 0$ we have
\begin{align*}
&| \, ||g_k(x) -h ||_\cV - ||g_l(x) - h||_\cV | \leq ||g_k(x) -g_l(x)||_\cV = \sum_{i\leq u} \langle g_k(x) - g_l(x), e_i \rangle_\cV =  \sum_{i\leq u} \langle g_k - g_l, \G_x e_i \rangle_\G , 
\end{align*}
where we used the Parseval equality and set $u$ as the dimension of $\cV$. Now, as $g_k(x)$ is weakly convergent we can find for each $i$ a
$N(i)$ such that $\langle g_k - g_l, K_x e_i \rangle_K < \epsilon/u$. Let $N = \max\{N(1),\ldots, N(u) \}$ then we have 
\begin{align*}
&| \, ||g_k(x) -h ||_\cV - ||g_l(x) - h||_\cV | < \epsilon.
\end{align*}
As $\R$ is complete the sequence $\{s_k(x,h)\}_{k\in \mathbb{N}}$ attains its limit for every pair $(x,h) \in \bigcup_{l \in \mathbb{N}} \bigcap_{k\geq l}\text{dom}(||g_k (x) - h ||_\cV)$, i.e. whenever the sequence is defined for a pair $(x,h)$. In particular, as $||g_k (x) - h ||_\cV$ is integrable there exists for every $k$ a measurable set $A_k \subset 
\text{dom}(||g_k (x) - h ||_\cV )$ with $P[A_k] = 1$. $s_k$ attains its limit on the set $D := \bigcup_{l \in \mathbb{N}} \bigcap_{k\geq l} A_k \in \Sigma$ and $P[D] =1$ (continuity of measure). 
Finally, $ ||g_k (x) - h ||_\cV \leq B \chi D + ||h||_\cV \leq (B + 1) \chi D + ||h||_\cV^2  =:f(x,h)$.  $f(x,h)$ is integrable as a sum of integrable functions. Hence, we can apply
the Lebesgue dominated convergence theorem \cite{FREM00}[Th. 123C] and we get that $\lim_{k\rightarrow \infty} ||g_k (x) - h ||_\cV$ is integrable and
$$ \lim_{k\rightarrow \infty} \Err[g_k ] = \int_{D} \lim_{k\rightarrow \infty} ||g_k (x) - h ||_\cV \, dP(x,h). $$
Furthermore,   $\lim_{k\rightarrow \infty} ||g_k (x) - h ||_\cV = ||\mu(x) -h ||_\cV$ on $D$ as for any $\epsilon > 0$ there exists a $N$ such that for all $k>N$
\begin{align*}
&| \, ||g_k (x) - h ||_\cV - ||\mu(x) -h ||_\cV |  \leq || g_k (x) - \mu(x) ||_\cV 
\leq \sum_{i \leq u} \langle g_k (x) - \mu(x),e_i \rangle_\cV = \sum_{i \leq u} 
\langle g_k  - \mu,\G_x e_i \rangle_\G < \epsilon
\end{align*}  
as $g_k $ converges weakly to $\mu$ and $u< \infty$. Hence, $\lim_{k\rightarrow \infty} \Err[g_k ] = \Err[\mu]$ and $\mu \in \cH_\G$. Also,
$$ \inf_{g \in \cH_\G} \Err[g] \leq \Err[\mu] \leq  \inf_{g \in \cH_\G} \Err[g] + 1/n$$
for all $n \in \mathbb{N}$. Therefore, $\inf_{g \in \cH_G} \Err[g] =\Err[\mu]$. Hence, $\mu^*:= \mu$ is a minimiser.
\end{proof}

Furthermore, the probability measure needs to factor as $P(h,x) = P_X(x) P(h|x)$ and 
there needs to be constants $c,d$ such that 
$$\int_{\cV} \left( e^{\frac{||h - \mu^*(x)||_{\cV}}{d}} - \frac{||h - \mu^*(x)||_{\cV}}{d} -1 \right) dP(h|x) \leq \frac{c^2}{2 d^2}.$$
In particular, $ \exp\Bigl(\frac{||h - \mu^*(x)||_{\cV}}{d}\Bigr) - \frac{||h - \mu^*(x)||_{\cV}}{d} -1 $ needs to be $P(\cdot|x)$  integrable.
In case that $P(h|x)$ is concentrated on $\cL := \{L(x,\cdot)| x \in \cX \}$ 
we have $ \exp\Bigl(\frac{||h - \mu^*(x)||_{\cV}}{d}\Bigr) - \frac{||h - \mu^*||_{\cV}}{d} -1 \leq \exp\Bigl(\frac{L(y,y) + B||\mu^*||_\G}{d}\Bigr) $ on $\cL$, where we used Corollary \ref{cor:RKHSDom}
and restricted $d >0$. As $\mu^*$ has finite RKHS norm and as we have a probability measure the assumption is always fulfilled in our case.

\subsection{Discussion of assumptions from \cite{DBLP:conf/icml/SongHSF09}} \label{app:SongAss}
We review conditions from \cite{DBLP:conf/icml/SongHSF09,FukSonGre10} that establish
the existence of $\mu\in\mathcal{H}_{\Gamma}$ such that $\Expect(h(Y)|X=x)=\left\langle h,\mu(x)\right\rangle _{\mathcal{H}_{L}}\:\forall h\in\mathcal{H}_{L}$,
as well as convergence results for an empirical estimate of $\mu$
(note that certain additional conditions over those stated in \cite{FukSonGre10}
are needed for the results to be rigorous, following the discussion
in \citep[Appendix, Theorem 6.6]{FukSonGre10}: we include these here). To describe
these conditions, some preliminary definitions will be needed.  Define
the uncentred covariance operators, \[
C_{XX}=\Expect_{X}K(X,\cdot)\otimes K(X,\cdot)\qquad C_{XY}=\Expect_{XY}K(X,\cdot)\otimes L(X,\cdot),\]
where the tensor product satisfies $(a\otimes b)c=\left\langle b,c\right\rangle _{\mathcal{H}_{K}}a$
for $a\in\mathcal{H}_{K}$ and $b,c\in\mathcal{H}_{L}$ (an analogous
result applies for $a,b,c\in\mathcal{H}_{K}$). As the mappings in
Song et al. are defined from $\mathcal{H}_{K}$ to $\mathcal{H}_{L}$,
rather than $\mathcal{X}$ to $\mathcal{H}_{L}$, we next establish
the relation between the Song et al. notation and ours: for every
$h\in\mathcal{H}_{L}$,\[
\left\langle h,\mu(x)\right\rangle _{\mathcal{H}_{L}}=\left\langle h,\mathcal{U}_{Y|X}\phi(x)\right\rangle _{\mathcal{H}_{L}},\]
where the notation on the left is from the present paper, that on
the right is from Song et al., and $\mathcal{U}_{Y|X}\::\:\mathcal{H}_{K}\rightarrow\mathcal{H}_{L}$
(we therefore emphasize $\mathcal{U}_{Y|X}\not\in\mathcal{H}_{\Gamma}$,
since its input space is $\mathcal{H}_{K}$, however $\mathcal{U}_{Y|X}\circ\phi(x)\in\mathcal{H}_{\Gamma}$
under the conditions stated below). We have the following theorem,
corresponding to \citep[Theorem 1]{DBLP:conf/icml/SongHSF09}.
\begin{theorem}[\cite{DBLP:conf/icml/SongHSF09}]
\label{thm:conditionalMeanEmbedConverge-1}Assum\emph{e that for all
$h\in\mathcal{H}_{L}$, we have }$\Expect\left(h(Y)|X=\cdot\right)\in\mathcal{H}_{K}$,
$C_{XX}$ is injective, and the operator $Sh:=\Expect\left(h(Y)|X=\cdot\right)$
is bounded for all $h\in\mathcal{H}_{L}$. Song et al. make the further
smoothness assumption that $C_{XX}^{-3/2}C_{XY}$ is Hilbert-Schmidt
(in fact, following \citep[Appendix]{FukSonGre10}, boundedness should
be sufficient, although this is outside the scope of the present paper).
Then the definition $\mathcal{U}_{Y|X}:=C_{YX}C_{XX}^{-1}$ satisfies
both $\mathcal{U}_{Y|X}\circ\phi(x)\in\mathcal{H}_{\Gamma}$ and $\mu(x)=\mathcal{U}_{Y|X}\phi(x)$,
and an empirical estimate $\widehat{\mathcal{U}}_{Y|X}:=\widehat{C}_{YX}\left(\widehat{C}_{XX}+\delta_{n}I\right)^{-1}$
converges with rate \[
\left\Vert \widehat{C}_{YX}\left(\widehat{C}_{XX}+\delta_{n}I\right)^{-1}\phi(x)-C_{YX}C_{XX}^{-1}\phi(x)\right\Vert _{\mathcal{G}}^{2}=O_{p}(n^{-1/4}).\]
when $\delta_{n}=n^{-1/4}$.
\end{theorem}

\section{Proofs} \label{sec:proofs}
\paragraph{Convergence result from \cite{DBLP:conf/icml/SongHSF09}}
The theorem in \cite{DBLP:conf/icml/SongHSF09} states the following:
Let $P$ be a probability measure and let $\hat U$ be the estimate of $U$ defined in \cite{DBLP:conf/icml/SongHSF09} then there exists a constant $C$ such that $\lim_{n \rightarrow \infty} P^n[ || U - \hat U||_{HS} > C n^{-1/8} ]  = 0,$ where HS denotes Hilbert-Schmidt norm. 



Under  
the assumption that $K(x,x) < B$ for all $x \in \cX$ this implies: 
for a probability measure $P$ and an estimate $\hat U$  of $U$
it holds that
there exists a constant $\tau$ such that $\lim_{n \rightarrow \infty} P^n[ \sup_{h \in \cB(\cV)} \Expect_X[ \Expect[h|X] -
 \langle h, \hat \mu(X) \rangle_\cV]^2  > \tau n^{-1/4} ] 
 = 0.$
\begin{proof}
Let $|||\cdot|||$ be the operator norm. We have that $|||A||| \leq ||A||_{HS}$ for
 a Hilbert-Schmidt operator \cite{WER02}[p.268,Satz VI.6.2c]. Using this fact and Cauchy-Schwarz we get:
\begin{align*}
&\sup_{h \in \cB(\cV)} \Expect_X[ \Expect[h|X] -
 \langle h, \hat U K(X,\cdot) \rangle_\cV]^2 \leq  \Expect_X
 ||(U - \hat U) \frac{K(X,\cdot)}{||K(X,\cdot)||_K}||_\cV^2 \, ||K(X,\cdot)||_K^2
 \leq |||U-\hat U|||^2 \, \Expect_X K^2(X,X) \\
 &\leq B ||U-\hat U||_{HS}^2
\end{align*}
and the statement follows with $\tau := C^2/B $.
\end{proof}

\paragraph{Example} 
The space $\cX$ together with the reproducing kernel of $\cH_K$ can lead to problems for the assumption from \cite{DBLP:conf/icml/SongHSF09}:
\begin{corollary}
Let $\cV$ be finite dimensional such that a function $\tilde h \in \cV$ exists
with $\tilde h(y) \geq \epsilon > 0$ for all $y \in \cY$. Furthermore, let
$\cX := [-1,1]$ and the reproducing kernel for $\cH_K$ be $K(x,y) = xy$. Then there
exists no measure which fulfills the assumption from eq. \ref{eq:SongAssumption}. 
\end{corollary}
\begin{proof}
Assume there is a measure and a measure space $(\cY,\Sigma,\mu)$. We can assume that all $ h \in \cV$ are integrable as otherwise the assumption fails.
We have $\Expect[\tilde h| X = x] > \Expect[\epsilon \chi \cY | X = x] = \epsilon 
> 0$. However, by assumption $\Expect[\tilde h| X = x] = \sum_{i=1}^m \langle \tilde h,h_i \rangle_L \Expect[\tilde h_i| X = x] =
\sum_{i\leq m } \langle \tilde h,  h_i\rangle_\cV  f_i(x)$ for some $f_i \in \cH_K$,
where $m$ is the dimension of $\cV$ and $\{h_i \}_{i \leq m}$ is an orthonormal basis. $\cH_K = \{z\cdot | z \in \mathbb{R} \}$ and hence $\Expect[\tilde h(Y)| X = x] = \langle \tilde h,   \sum_{i\leq m} z_i x h_i \rangle_\cV = x \langle \tilde h,   \sum_{i\leq m} z_i  h_i \rangle_\cV$. If the inner product is positive then we get  $\Expect[\tilde h(Y)| X = x] \leq 0$ for $x \in [-1,0]$ and a contradiction. Similarly for the case where the inner product is negative.
\end{proof}


\section{Justification of sparsity conditional mean embedding}
\label{app:massi}
In this section, we provide a justification for the objective function (16) which we used in Section 5 
to derive a sparse conditional mean embedding. Specifically, 
we show that the objective function (16) provides a natural upper bound to (5), which is the error measure 
we used to derive the embedding itself.

For simplicity, we assume that the underlying conditional mean embedding belongs to a reproducing kernel Hilbert space of ${\cal H}_L$-valued functions, ${\cal H}_\Gamma$, with operator-value kernel $\Gamma \in {\cal L}({\cal H}_L)$. That is, there exists $\mu^* \in {\cal H}_\Gamma$, such that, for every $h \in {\cal H}_L$ and $x \in {\cal X}$, it holds that $\E_Y [h(Y)|X=x] = \langle \mu^*(x),h\rangle_L$. 
Under this assumption we have that
\begin{eqnarray}
\nonumber
\E_Y[h(Y)|X] - \langle h,\mu(X)\rangle_L 
& = & \langle \mu^*(x) - \mu(x),h\rangle_L\\ \nonumber
& = & \langle \mu^* - \mu,\Gamma(x,\cdot) h\rangle_\Gamma \\ \nonumber
& \leq & \|\mu^* - \mu\|_\Gamma \|\Gamma(x,\cdot)h\|_L \\ \nonumber
& \leq &
\|\mu^* - \mu\|_\Gamma \||\Gamma(x,x)\||^{\frac{1}{2}} \|h\|_L.
\nonumber
\end{eqnarray}
where the second equality follows from the reproducing property of the kernel $\G$, the first inequality follows from Cauchy Schwarz's inequality,  the last inequality follows 
from \citep[Proposition~1(d)]{MIC05} and $\||\Gamma(x,x)\||$ denotes the operator norm. Consequently,
\begin{eqnarray}
\nonumber
{\cal E}[\mu] & = & \sup_{\|h\|_L \leq 1} \E_X \left[\left(\E_Y[h(Y)|X] - \langle h,\mu(X)\rangle_L\right)^2\right] \\
& \leq & c \|\mu^* - \mu\|_\Gamma^2 
\label{eq:gino}
\end{eqnarray}
where we have defined $c = \E_X \||\Gamma(X,X)\||$.

We choose $\mu$ to be the solution ${\hat \mu}$ of problem (9). As noted at the beginning of Section 5, in many practical circumstances, we wish to approximate ${\hat \mu}$ by a sparse version 
$\mu_{\rm sparse}$ which involves a smaller number of parameters. For this purpose, it is natural 
to use the error measure $\|\cdot\|_{\Gamma}$. Indeed, equation \eqref{eq:gino} and 
the triangle inequality yield that 
$$
{\cal E}[\mu_{\rm sparse}] \leq c \left(\|\mu^* - {\hat \mu}\|_\Gamma + \|{\hat \mu}-\mu_{\rm sparse}\|_\Gamma\right)^2.
$$
Thus if ${\hat \mu}$ estimates well the true embedding $\mu^*$, so will the sparse mean embedding $\mu_{\rm sparse}$.

In the specific case considered in Section 5, $\Gamma(x,x') = K(x,x') Id$, so that 
$\||\Gamma(x,x)\||= K(x,x)$ and $\|\cdot\|_\Gamma = \|\cdot\|_{K \otimes L}$. Furthermore, we choose 
${\hat \mu} = \sum_{i,j=1}^n W_{ij} K_{x_i} L_{y_j}$ and $\mu_{\rm sparse} =  \sum_{i,j=1}^n M_{ij} K_{x_i} L_{y_j}$, where matrix $M$ is encouraged to be sparse. Different sparsity methods are discussed in Section 5, such as the the Lasso method (cf. equation (15)).

\end{onecolumn}

\end{document}